# Music Style Classification with Compared Methods in XGB and BPNN

Lifeng Tan, Cong Jin*, Zhiyuan Cheng
School of Information and Communication Engineering,
Communication University of China
Beijing, China
Correspondence: jincong0623@cuc.edu.cn

Xin Lv, Leiyu Song
School of Animation and Digital Arts,
Communication University of China
Beijing, China
lvxincuc@163.com

*Abstract*—Scientists have used many different classification methods to solve the problem of music classification. But the efficiency of each classification is different. In this paper, we propose two compared methods on the task of music style classification. More specifically, feature extraction for representing timbral texture, rhythmic content and pitch content are proposed. Comparative evaluations on performances of two classifiers were conducted for music classification with different composers' styles. The result shows that XGB (eXtreme Gradient Boosting) is better suited for small datasets than BPNN (Back Propagation Neural Network).

## I. Introduction

People can now download more and more songs freely from the Internet. Consequently, the collection and arrangement of music have become an important subject. Automatic classification of unidentified music is the focus of this paper. Choosing the right features for classification is a difficult problem. After consulting the literature and opinions of experts, we selected eight melody-related features. Features are extracted by using jSymbolic, which is particularly good at processing music in MIDI format. We mainly use two machine learning methods, XGB (eXtreme Gradient Boosting) and BPNN (Back Propagation Neural Network) to classify music according to composers. We use some evaluation metrics, especially AUC (Area Under roc Curve), to assess their performance.

The rest of the paper is arranged as follows. Section II introduces the existing approaches applied to music classification. Section III briefly describes how to extract features and eight feature parameters. The principle and brief deduction of two classifiers are also shown in Section III. Section IV involves in data sets and experimental settings. The evaluation index of classifiers and the results are reported in Section V. The conclusions are in the Section VI.

## II. Literature Review

Music classification is one of the hotspots of computer scientists in recent years. Music can be categorized according to different characteristics. Currently studies mainly focus on two aspects: one is classified according to the genre[4], the other is classified according to emotion. Traditional music classification methods are based on supervised machine learning (Tzanetakis and Cook, 2002). They used KNN (k-Nearest Neighbor) and Gaussin Mixture model[3]. The above methods as well as MFCC (Mel-Frequency Cepstral Coefficients) were used for noisy classification[5]. However, these methods have difficulties in directly applying to multi-classification problems. Huang and his team (2012) built a new algorithm, ELM (Extreme Learning Machine)[1]. It had better generalization performance.

Because the application of CNN (Convolution Neural Network) in image classification is very successful, it has also been used in music classification recently[8]. Md. Shamim Hussain and Mohammad Ariful Haque (2018) developed SwishNet—— a fast CNN for audio data classification and segmentation[7]. AzarNet, a DNN (Deep Neural Network) was created by Shahla Rezezadeh Azar and his colleagues (2018) to recognize classical music. Liu and Feng (2019) fully took advantages of low-level information from spectrograms of audios and developed a new CNN algorithm[6].

## III. Methodology

### A. Features Extraction

In this paper, we mainly use jSymbolic to extract features. jSymbolic is a useful tool to extract features that we need for classification easily. It is an easy-to-learn software, whose user only need to have a basic understanding of Java[9]. Besides, it is particularly good at processing music in MIDI format. So we just need to import the music in MIDI format into jSymbolic and extract the desired features through a series of operations.

### B. Feature Parameters

In order to solve the deficiency of single feature, this paper extracts eight musical features. These features mainly describe the characteristics of melody itself. Therefore, a lot of music-related terms are involved in this part.

- Range. This parameter represents differences in semi tone between the highest and lowest pitches. A semi tone is the smallest distance between two pitches. A black key on the piano represents a semi tone.

- Repeated Notes. This parameter shows how many melody tones are repeated notes. Melody tone, a term in the field of music, refers to two notes played successively.

- Vertical Perfect Fourths. This parameter represents the vertical intervals between two meters. Meter is the unit that measures rhythm in the field of music. Fourths means that a meter consists of quarter notes and there is one beat each meter.
- Rhythmic Variability. This parameter represents the variations based on the original tune. Some notes of music are played many times, which are called as theme or main melody. There are some changes in the process of repetition.
- Parallel Motion. This parameter represents how many parallel motion in a song. Upward motion refers to pitches from lower to higher. As contrast, downward motion refers to pitches from higher to lower. It is parallel motion when is neither upward motion or downward motion.
- Vertical Tritones. This parameter represents the vertical interval between two tritones. The tritone is a melody tone consisted of three successive tones.
- Chord Duration. This parameter is the average of chord duration. Chord refers to a group of notes with a certain relationship. Chords crated by different composers are different. Duration of some are short, while others are longer.
- Number of Pitches. This parameter represents a total number of notes, especially including rests. Rest is a mark recording the duration of a pause in music.

*C. Classifiers*

*1) eXtreme Gradient Boosting*

XGB, proposed by Dr.chen in 2016, is a large-scale machine learning method for tree boosting and the optimization of GBDT(Gradient Boosting Decision Tree). As a lot of researches have mentioned, GBDT is an ensemble learning algorithm, which aims to achieve accurate classifications by combining a number of iterative computation of weak classifiers(such as decision trees). However, unlike GBDT, XGB can take advantages of multi-threaded parallel computing by using CPU automatically to shorten the process of iteration. Besides, additional regularization terms help decrease the complexity of the model.

In Supervised Learning, there are objective function as well as predictive function. In XGB, objective function in Equation (1) consists of training loss $L(\phi)$ which measures whether model is fit on training data and regularization $\Omega(\phi)$ which measures complexity of model[2]. If there is no regularization or regularization parameter is zero, the model returns to the traditional gradient tree boosting.

$$Obj(\phi) = L(\phi) + \Omega(\phi) \quad (1)$$

When it comes to $L(\phi)$:

$$L(\phi) = \sum_{i=1}^{n}(y_i, \dot{y}_l) \quad (2)$$

$y_i$ is true value and $\dot{y}_l$ is predicted value.

If a model after an iteration is:

$$\dot{y}_l = \sum_{m=1}^{M} f_m(x_i), f_m \in A \quad (3)$$

Then the corresponding objective function is:

$$Obj(\phi) = \sum_{i=1}^{n} l(y_i, \dot{y}_l) + \sum_{m=1}^{M} \Omega(f_m) \quad (4)$$

And the model after t times iteration:

$$Obj^{(t)} = \sum_{i}^{n} l(y_i, \dot{y}_l) + \sum_{i=1}^{t} \Omega(f_i)$$
$$= \sum_{i=1}^{n} l(y_i, \dot{y}_l^{(t-1)} + f_t(x_i)) + \sum_{i=1}^{t} \Omega(f_i) \quad (5)$$

$f_t(x_i)$ is a predictable function newly added in the t times iteration.

The formula of second-order Taylor expansion is:

$$f(x + \Delta x) \approx f(x) + f'(x)\Delta x + \frac{1}{2} f''(x)\Delta x^2 \quad (6)$$

Using the second-order Taylor expansion of

$l(y_i, \dot{y}_l^{(t-1)} + f_t(x_i))$:

$$obj^{(t)} = \sum_{i=1}^{n} [l(y_i, \dot{y}_i^{(t-1)}) + g_i f_t(x_i) + \frac{1}{2} h_i f_t^2(x_i)] + \sum_{i=1}^{t} \Omega(f_i) \quad (7)$$

In this formula, $g_i = \dfrac{\partial l(y_i, \dot{y}_l^{(t-1)})}{\partial \dot{y}_l^{(t-1)}}$ is the first-order derivative of $l(y_i, \dot{y}_l^{(t-1)})$. $h_i = \dfrac{\partial^2 l(y_i, \dot{y}_l^{(t-1)})}{(\partial \dot{y}_l^{(t-1)})^2}$ is the second-order derivative of $l(y_i, \dot{y}_l^{(t-1)})$.

When it comes to $\Omega(f)$:

$$f_t(x_i) = \omega_{q(x_i)} \quad (8)$$

$$\Omega(f_t) = \gamma T + \frac{1}{2} \lambda \sum_{j=1}^{T} \omega_j^2 \quad (9)$$

In Equation (8), $q(x_i)$ structure function, which describes the structure of a decision tree $\omega$ is the weight of the leaves on the tree. Equation (9) describes the complexity of a tree. $\gamma$ is a coefficient of leaf nodes, taking pre-processing to prune leaves while optimizing the objective function. $\lambda$ is another coefficient to prevent the model from over-fitting.

Define $P_j = \{i | q(x_i) = j\}$ as the sample set for each leaf $j$. Then objective function can be simplified as:

$$Obj^{(t)} = \gamma T + \sum_{j=1}^{T}[(\sum_{i \in P_j} g_i)\omega_j + \frac{1}{2}(\sum_{i \in P_j} h_i + \lambda)\omega_j^2] \quad (10)$$

When structures of trees $q$ are known. This function has solutions:

$$\omega_j^* = -\frac{\sum_{i \in P_j} g_i}{\sum_{i \in P_j} h_i + \lambda} \quad (11)$$

$$Obj^* = \gamma T + \frac{1}{2}\sum_{j=1}^{T}\left(\frac{\sum_{i \in P_j} g_i}{\sum_{i \in P_j} h_i + \lambda}\right)^2 \quad (12)$$

Blessed with traits mentioned above, XGB has the following advantages compared to traditional method. a) Avoiding over-fitting. According to Bias-variance trade-off, the regularization term simplifies the model. Simpler models tends to have smaller variance, thus avoiding over-fitting as well as improving accuracy of the solution. b) Supporting for Parallelism. Before training, XGB sorts the data in advance, and saves it as a block structure. When splitting nodes, we can calculate the greatest gain of each feature with multi-threading by using this block structure. c) Flexibility. XGB supports user-defined objective function and evaluation function as long as the objective function is second-order derivable. d) Built-in cross validation. XGB allows cross validation in each round of iterations. Therefore, the optimal number of iterations can be easily obtained. e) Process of missing feature values. For a sample with missing feature values, XGB can automatically learn its splitting direction.

*2) Back Propagation Neural Network*

BPNN is a multi-layer feedforward neural network trained by error back propagation learning algorithm. It was firstly coined by Rumelhart and McClelland in 1986. Blessed with Strong ability of non-linear mapping, generalization and fault tolerance, it has become one of the most widely used neural network models. The core of BP neural network mainly includes two parts: the forward propagation of signals as well as the reverse propagation of errors. In the former, input signals as input cells activate the cells of hidden layer and transfer information to them with weights. The hidden layer also acts on the output layer in this way, thus finally getting the output results. If those results are not fit on the expected output results, it turns to the latter process. The output layer error will be back-propagated layer by layer. The weights of the network are adjusted at the same time to make the output of the forward propagation process closer to the ideal output.

*a) the Forward Propagation*

Firstly, we need to introduce activation functions. It helps to solve complex non-linear Problems. The widely used activation function is the $sigmoid(x)$.

$$sigmoid(x) = \frac{1}{1+e^{-x}} \quad (13)$$

$$sigmoid'(x) = sigmoid(x)[1-sigmoid(x)] \quad (14)$$

In input layer, the input and output of the $i^{th}$ cell are the same. And the number of input cells is $n1$.

$$O_i = I_i \quad (15)$$

In hidden layer, the input and output of the $i^{th}$ cell are as follow. And the number of input cells is $n2$

$$I_j = \sum_{i=1}^{n1} \omega_{ji} O_i + \delta_j \quad (16)$$

$$O_j = sigmoid(I_j) \quad (17)$$

$\omega_{ji}$ is the weight connecting the $i^{th}$ input cell and the $i^{th}$ hidden cell. $\delta_j$ represents the thresholds of the $j^{th}$ hidden cell.

In output layer, the input and output of the $k^{th}$ cell are as follow.

$$I_k = \sum_{j=1}^{n2} \omega_{kj} O_j + \delta_k \quad (18)$$

$$O_k = sigmoid(I_k) \quad (19)$$

$\omega_{kj}$ is the weight connecting the $j^{th}$ hidden cell and the output cell. $\delta_k$ represents the thresholds of the output cell.

*b) the Back Propagation*

We define the expected output is $\hat{O}$ and the number of the output cells is $n3$. After training, the total error is:

$$E = \frac{1}{2}\sum_{k=1}^{n3}(\hat{O}_k - O_k)^2 \quad (20)$$

According to the chain rule, we can adjust the weight.

$$\frac{\partial E}{\partial \omega_{ji}} = \frac{\partial E}{\partial O_k} \cdot \frac{\partial O_k}{\partial I_k} \cdot \frac{\partial I_k}{\partial \omega_{kj}} = (\hat{O}_k - O_k) \cdot O_k \cdot (1-O_k) \cdot O_j \quad (21)$$

Similarly, other weights can be adjusted in this way.

## IV. EXPERIMENT

### A. Data sets

As for data sets, we selected classical music from five composers. They are Haydn, Mozart, Beethoven, Bach and Schubert. We get 200 pieces of music from each composer, a total of 800 pieces. We use 90% of each composer's data as a training set and the remaining 10% is used as testing set. Music pieces are all in MIDI format.

### B. Experimental Settings

In this experiment, we employ python 3, an interpretative scripting language. We mainly use sklearn toolbox to deal with data, which is simple but efficient tools for data mining and data analysis. It is not only accessible to beginners but also reusable in various contexts. Matplotlib toolbox is used to draw the ROC curve.

## V. EVALUATION

It is significant to evaluate the performance of classifiers. Here we use some evaluation metrics to achieve this aim.

### A. Evaluation Metrics

- Accuracy. Accuracy refers to the percentages that samples correctly classified accounts for the whole samples.

$$A = \frac{N_{tp}}{N_{tp} + N_{fp} + N_{fn}} \quad (22)$$

- Precision. Precision refers to the percentages that positive samples correctly classified accounts for the whole positive samples predicted by classifiers.

$$P = \frac{N_{tp}}{N_{tp} + N_{fp}} \quad (23)$$

- Recall. Recall refers to the percentages that positive samples correctly classified accounts for the whole actual positive samples. It is also called as sensitive.

$$R = \frac{N_{tp}}{N_{tp} + N_{fn}} \quad (24)$$

- F-Measure. F-Measure refers to harmonic weight mean of precision and recall. It synthesizes the results of precision and recall. Therefore, the method is more effective when f-measure is higher.

$$F = \frac{2PR}{P + R} \quad (25)$$

- Receiver Operating Characteristic. In ROC space, the abscissa is FPR (false positive rate and the ordinate is TPR (true positive rate). It describes a trade-off between them.

- Area Under roc Curve. As the name suggests, the value of AUC is the area below the ROC curve. Usually, AUC values range from 0.5 to 1.0, and larger values represent better performance.

### B. Results

The values of two classifiers have been placed in TABLE I. As we can see, there is no notable difference in AUC between XGB and BPNN, with 0.55 and 0.54 representatively. However, in other four values, XGB obviously performs better than BPNN. Precision and Recall of XGB are 0.31 and 0.41 representatively. They are 0.18 and 0.26 in BPNN. Accuracy of XGB is 0.63, while the figure of BPNN only is 0.43. F-Measure of XGB is the average of Precision and Recall, at 0.35. F-Measure of BPNN is low, with a mere 0.22. XGB is better at dealing with small-scale data sets.

TABLE I. PERFORMANCE COMPARISONS ON CLASSIFIERS

| Classifiers | Accuracy | Precision | Recall | F-Measure | AUC |
|---|---|---|---|---|---|
| XGB | 0.63 | 0.31 | 0.40 | 0.35 | 0.55 |
| BPNN | 0.43 | 0.18 | 0.26 | 0.22 | 0.54 |

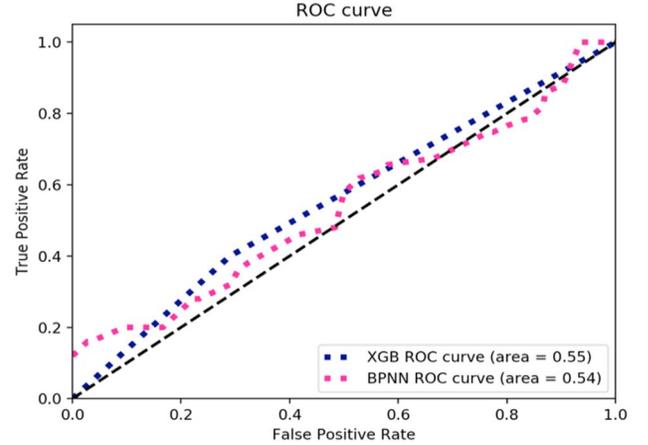

Fig. 1. ROC curve of XGB and BPNN

The performance of XGB and BPNN is shown in Figure1. ROC curve of BPNN has more twists and turns, compared with that of XGB. The highest point of ROC curve of XGB is closer to the upper left corner, which means the results of XGB are more accurate.

## VI. CONCLUSION

This paper mainly compares the performance of two classifiers for music style classification with different composers. Firstly, we choose eight features according to expert opinion and documentation. The next step is to use jSymbolic to extract features. We use the data to train two classifiers and use evaluation metrics to assess their behaviors. As a result, XGB outperforms BPNN. We also show XGB has higher accuracy as well as F-measure than BPNN although they have similar AUC.

ACKNOWLEDGMENT

This paper is supported by "the Fundamental Research Funds for the Central Universities".